\documentclass[runningheads]{llncs}
\usepackage{svg}
\usepackage{graphicx}
\usepackage{tabularx}
\usepackage{caption}
\usepackage[frozencache,cachedir=.,outputdir=.]{minted}
\begin{document}
%
% \title{The Director: A Tree-Based Software Behaviour Framework and Algorithm with Soft Transitions}
% \title{The Director: A Behaviour Tree Framework and Algorithm}
\title{The Director: A Composable Behaviour System with Soft Transitions}
%
%\titlerunning{Abbreviated paper title}
% If the paper title is too long for the running head, you can set
% an abbreviated paper title here
%
\author{Ysobel Sims \and Trent Houliston \and Thomas O'Brien \and Alexandre Mendes \and Stephan Chalup}
\authorrunning{Y. Sims et al.}
% First names are abbreviated in the running head.
% If there are more than two authors, 'et al.' is used.
%
\institute{University of Newcastle, Callaghan 2308, Australia}
\maketitle              % typeset the header of the contribution
\begin{abstract}

Software frameworks for behaviour are critical in robotics as they enable the correct and efficient execution of functions. While modern behaviour systems have improved their composability, they do not focus on smooth transitions and often lack functionality. In this work, we present the Director, a novel behaviour framework that addresses these problems. It has functionality for soft transitions, multiple implementations of the same action chosen based on conditionals, and strict resource control. The system was successfully used in the 2022/2023 Virtual Season and RoboCup 2023 Bordeaux, in the Humanoid Kid Size League. It is implemented at \textit{\url{https://github.com/NUbots/DirectorSoccer}}, which also contains over thirty automated tests and technical documentation on its implementation in NUClear.

\end{abstract}
\section{Introduction}

Autonomous robotics is a broad field where robots perform diverse tasks. The actions performed at any given moment depend on state and environment information. Behaviour frameworks allow developers to define rules for behaviour algorithms to manage a robot's actions. There are many desirable features for a behaviour system:

\begin{itemize}
    \item[$\bullet$] It should facilitate a smooth transition between actions.
    \item[$\bullet$] It should account for differences over time in information quality and environment knowledge.
    \item[$\bullet$] It should be flexible and versatile.
    \item[$\bullet$] Modules competing for resources should not be able to use a resource at the same time, such as motors.
    \item[$\bullet$] The behaviour should be composable for quick development.
    \item[$\bullet$] State information should be available for debugging.
    \item[$\bullet$] The system should run in real time.
\end{itemize}

Looking back at previous research on behaviour systems, the classical subsumption system~\cite{Brooks1986} provided a modular approach but lacked composability and functionality. Behaviour trees are modular, reactive and composable but lack complex functionality and can become cumbersome~\cite{biggar2021}. They cannot change actions based on the existence of environment knowledge, which research aims to address~\cite{Aisu1995,Yang2021,Safronov2020}. The $ABC^2$ behaviour system previously used in RoboCup used a queue-based agenda focusing on multi-agent gameplay. It did not facilitate smooth transitions and lacked complexity beyond multi-agent functionality. The Dynamic Stack Decider (DSD)~\cite{Poppinga2022} is composable and maintainable but lacks the desired functionality for soft transitions. The Humanoid Control Module~\cite{Bestmann2020} was incorporated with the DSD to abstract lower-level hardware modules from higher-level strategy modules and to avoid conflicts in motor control. It suggests that the DSD alone does not provide all desired functionality for a complete behaviour framework. 

The literature has improved over time towards composable, functional and transparent systems. However, each still lacks some desired functionality. Research often focuses on one or a few key components, such as fidelity of information~\cite{Veloso2000}, without considering all aspects needed for a general behaviour system. This research aims to address these gaps.

Our framework leverages the popular publisher/subscriber pattern~\cite{Eugster2003} found in many message-passing systems. In this pattern, publishers distribute messages to subscribers in a loosely coupled, asynchronous manner. This approach is used by systems such as Robot Operating System (ROS)~\cite{Quigley2009} and NUClear~\cite{Houliston2016}.

In this work, we present The Director, a behaviour framework for autonomous systems that emphasises modularity and transitions. It incorporates functionality for soft transitions to ensure safe robot motions as tasks change. Its modular architecture enables programmers to focus on developing one small and specific functionality for the robot at a time. The framework's versatility facilitates complex behaviours, including defining multiple ways to complete one task chosen based on conditionals. A library of automated tests~\footnote{\scriptsize{https://github.com/NUbots/DirectorSoccer/tree/main/module/extension/Director/tests}} supports the creation of the Director's backend algorithm. The Director's design integrates many desirable features into a simple, coherent framework.

The Director exists at a high level of abstraction that oversees all software levels of the system. It imposes strict resource control and facilitates transitions. The Director can work together with algorithms that provide the specific logic for how a robotic agent will achieve its goal, such as large language models (LLMs)~\cite{OBrien2023,ProgPrompt} and knowledge-base systems~\cite{KnowRob,KnowRob2}, by providing the framework for these systems to work within. 

The Director is general-purpose and implementable in reactive message-passing robotic software architectures. Our team implemented it within NUClear for the RoboCup Humanoid Soccer League, shown in both the 2022/2023 Virtual Season and RoboCup 2023 Bordeaux, with a game won using the Director in each competition. We successfully converted from a subsumption-based system to the Director and have exclusively used this system since the 2022/2023 Virtual Season. The Director framework simplified the implementation of new behaviours, provided more flexibility and facilitated published work in using LLMs to power the higher level reasoning of a robot~\cite{OBrien2023}. 

\section{Framework}\label{sec:algo}

The Director is a framework for controlling the flow of behaviour in a system. A fundamental aspect of the framework is the concept of Providers and Tasks which build a tree structure, where Tasks are messages requesting functionality, and Providers provide the functionality for those behaviours. Providers can call subtasks, which are considered children of the Provider. The Director has a wide range of features to solve limitations in existing behaviour systems. Definitions of key terms are in Table~\ref{tab:definitions}.

\begin{figure}
    \centering
    \includegraphics[width=\textwidth]{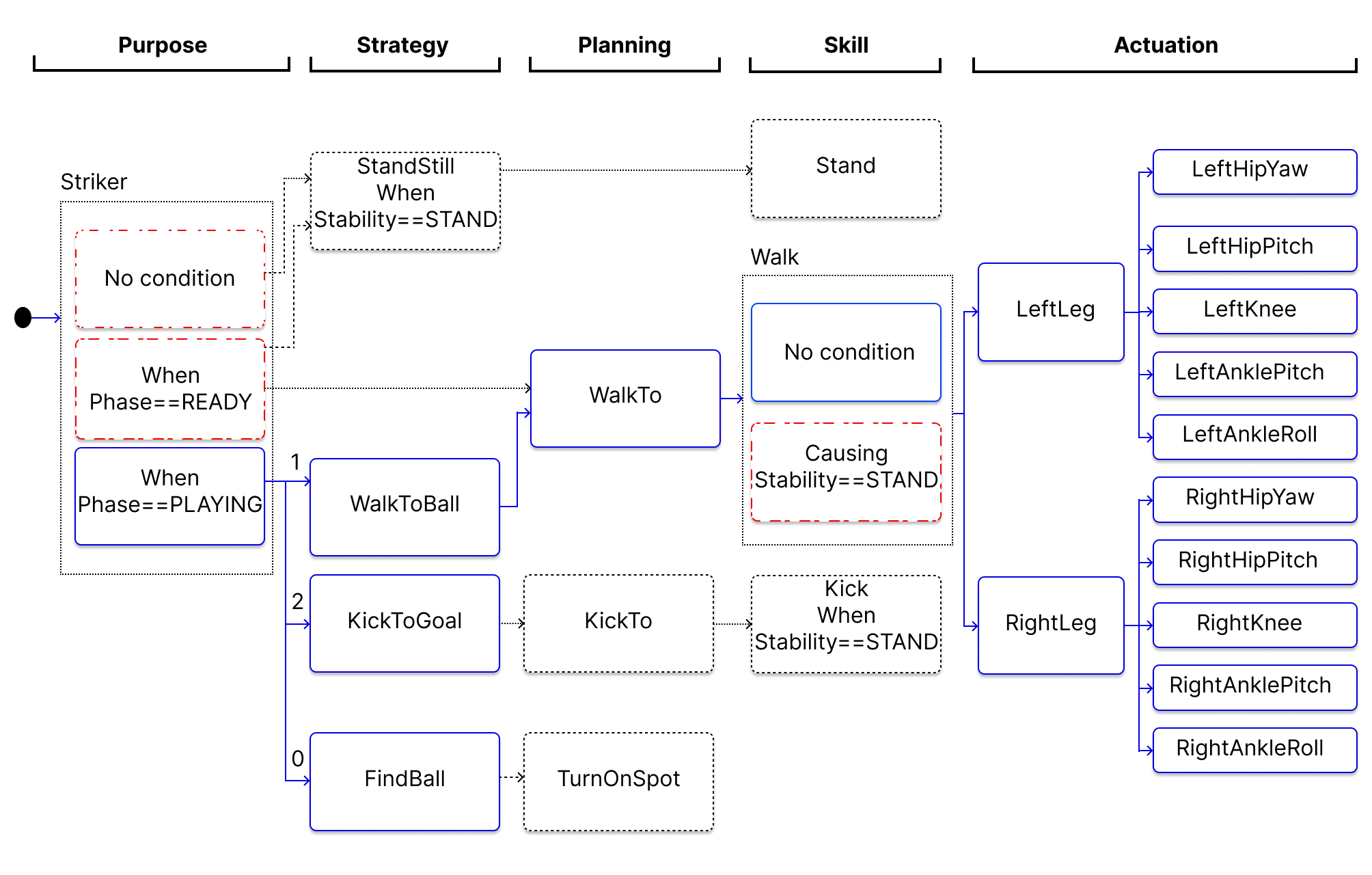}
    \caption{This figure shows a Director graph for a humanoid robot playing soccer. It shows active (solid blue), blocked (dashed red), and inactive (dotted black) Providers and Tasks. The values of Tasks indicate their priority.}
    \label{fig:system}
    \vspace{-2.5em}
\end{figure}

We use a soccer-playing scenario to describe the concepts in the Director, where a humanoid robot approaches a ball and kicks it. The transition from walking to kicking should be stable. Figure~\ref{fig:system} shows the Director graph at the start of this scenario, where the robot has seen the ball and is walking towards it.

\begin{table}
    \centering
    % \begin{tabular}{|p{0.2\textwidth} | p{0.8\textwidth}|}
     \begin{tabular}{|c|c|}
        \hline
        \textbf{Concept} & \textbf{Description} \\ \hline
        Provider & A function that provides the functionality for a Task. \\ \hline
        Non-Provider & A function that doesn't provide for a Task. \\ \hline 
        Provider Group & A collection of Providers that service the same Task. \\ \hline 
        Task & A request for a single piece of functionality. \\ \hline
        Subtask & Task from a Provider. \\ \hline 
        Root Task & A Task from a non-Provider. \\ \hline 
        Priority & A value that determines if a Task can take control. \\ \hline  
        Optional & The Task is not required to run and will defer to non-optional Tasks. \\ \hline 
        Done & A Task that indicates that the Provider has completed its Task. \\ \hline 
        Idle & A Task that will keep running the last set of Tasks for a Provider. \\ \hline      
    \end{tabular}
    \vspace{6.0pt}
    \caption{Definitions for key terms for the Director framework.}
    \label{tab:definitions}
    \vspace{-1.2em}
\end{table} 

\begin{table}
    \centering
    % \begin{tabular}{|p{0.24\textwidth} | p{0.73\textwidth}|}
    \begin{tabular}{|c|c|}
        \hline
        \textbf{DSL Keywords} & \textbf{Description} \\ \hline
        Provide & The normal Provider that provides the functionality for a Task. \\ \hline 
        Start & The Provider that is run when a Provider group gains control. \\ \hline 
        Stop & The Provider that is run when a Provider group loses control. \\ \hline 
        Needs & The Provider needs control of the specified subtask to run. \\ \hline 
        When & Conditions on when the Provider can be active. \\ \hline 
        Causing & The state that will be achieved by running this Provider. \\ \hline 
        Uses & Retrieves information on a Provider's subtask state. \\ \hline 
        RunReason & Gives information on why the Provider ran. \\ \hline         
    \end{tabular}
    \vspace{6.0pt}
    \caption{DSL keywords used in the Director.}
    \label{tab:dsldefs}
    \vspace{-2.0em}
\end{table}

A behaviour system can be split into five main sections, as shown at the top of Figure~\ref{fig:system}. The Director framework does not have any knowledge of these layers, but they can be optionally incorporated into the program's structure to further conceptualise and modularise the behaviour. The literature uses different terms interchangeably, which can lead to confusion. 

We propose the following terminology. The actuation layer involves controlling the robot's hardware, such as motors and audio output. The skill layer is responsible for performing complex physical movements of the robot. The planning layer determines when and in what way the skill layer is called based on the current state of the environment. The strategy layer makes specific high-level decisions by stringing together planners that run based on priority and state information. Strategy modules are often small and do not receive specific data from the layer above. The purpose layer is at the top of the Director tree and determines the overall goal of the robot.

The skill and actuation layers can abstract platform-specific functionality from the higher-level behaviour layers. The Humanoid Control Module introduced by Bestmann et al.~\cite{Bestmann2020} abstracts the robot's motions from the higher behaviour layers and avoids conflicts in motor control. They focus on using the same higher-level modules across legged and wheeled robots, with the lower-level control module implementing the platform-specific skills and actions. The Director inherently facilitates this abstraction through loose coupling. 

The remainder of this section will describe in detail the concepts that form the Director framework. By implementing these together, developers can take advantage of the benefits the Director inherently brings.

\subsection{Providers}

Providers are functions that perform actions to satisfy the requirements of one particular Task, either directly or by running subtasks. The function runs when the Task it provides for is requested from anywhere in the system. Providers may have conditions on when they can run, defined in their declaration. Given these conditions are satisfied, the Provider will run when a new task is received. 

In our soccer-playing scenario, the boxes in Figure~\ref{fig:system} are Providers. In the beginning, the robot should walk to the ball. The WalkToBall box is a Provider that provides the functionality for the WalkToBall Task. It finds the ball position and requests a WalkTo Task with this ball position data. The WalkTo Task manages what velocity to send to the walk engine based on the target position.

\subsubsection{Provider Groups:}

A group of Providers for one Task type is called a Provider group. Only one Provider in a group can run at a time, with the running Provider chosen based on the conditions of the Providers. If a Provider can no longer run, the Task is reassigned to another Provider in the group. Subtasks from a Provider in a group are from the group, not the specific Provider.

Striker and Walk are Provider groups in Figure~\ref{fig:system}. The Striker is using the `playing' Provider since that condition is true. If the game phase were to transition to the ready state, the `ready' Provider would become active. Provider groups allow for a change in behaviour for a specific Task when there is a change in environment or state information. This flexibility is a desired trait explored in the literature~\cite{Veloso2000,Safronov2020,Yang2021}.

\subsubsection{Provider Types:}

Provider groups facilitate to inclusion of three unique Provider types. The standard type executes functionality for a Task, as discussed previously. 

The `start' type sets up the state of the Provider when a Provider group gains control of its Task after not having control previously. It runs once and then the standard Provider will run.

The `stop' type is used to clean up code and only runs once when a Provider group is losing control of its Task. It cannot run without a start or standard type Provider running before it. 

In our scenario, the Walk Provider group will have a start Provider that sets up the walk engine. When the Kick Provider takes over, the Walk Provider group will have a stop Provider that cleans up the walk engine.

\subsubsection{Needs:}

The DSL keyword `Needs' is used to ensure that a Provider can only run if it has priority to take control of its subtasks. 

In our scenario, the skill Providers Need the relevant limbs for their Task. For example, the Walk and Kick Providers will specify that they Need the LeftLeg and RightLeg Tasks. This conflict means the Walk and Kick Providers cannot run simultaneously. Because the KickToGoal Task has higher priority than the WalkToBall Task, if both are requested to run then the Kick will take control and the Walk will be blocked from running as it does not have control of its Needs subtasks.

\subsubsection{When:}

`When' is a DSL keyword that only allows a Provider to execute when the system satisfies a specified condition. The condition has three parts - a state type, a comparison operator and a state value. 

In our scenario, the Kick only runs When the stability state of the robot is equal to standing. This prevents the robot from transitioning from the walk mid-step and falling over. 

This functionality, combined with Provider groups, solves the problem described by Veloso et al.~\cite{Veloso2000}, where existing systems don't consider the precision of environment knowledge. They proposed a behaviour system that emphasised the accuracy and fidelity of information in determining how to perform a task. A single action, such as localising, could have multiple methods for execution depending on the quality of sensor data. In the Director, each method can have an associated Provider, with the precision level for that method defined in the When conditional.

\subsubsection{Causing:}

The DSL keyword `Causing' in the Director allows for soft transitions between Providers. It declares that the Provider will cause a condition to be satisfied. Similarly to `When', it has a state type, a comparison operator and a state value. If a Provider has a `When' condition, the Director will prioritise lower priority Providers with a matching `Causing' if the `When' fails.

In our scenario, the Kick requires the robot to be standing, but the robot is currently walking. A version of the Walk in the Walk Provider group causes the stability state to become equal to standing. Because the Kick has a higher priority, this will run and make the walk stop cleanly and stand still, allowing the Kick to execute safely.

\subsection{Tasks}

Tasks are jobs that Providers execute and can contain data. They have three pieces of information: Task data, priority level, and an optional flag. Prioritisation of Tasks determines which should run, with non-optional Tasks always taking priority over optional ones. If a Task is optional, other subtasks can run if it cannot, but this is not true for non-optional subtasks. The Director will implement all-or-nothing for non-optional subtasks. Both Providers and non-Providers can request Tasks.

In our scenario, the Walk Provider requests LeftLeg and RightLeg subtasks. The Provider must have control over both Tasks for them to run. The Walk Provider can also have optional LeftArm and RightArm subtasks, which will run if possible but will not block the other subtasks from running.

\subsubsection{Root Tasks:}

Tasks requested from non-Providers are called root tasks and are the starting points of the Director graph. These tasks are siblings at the top of the tree. Root tasks are different from Tasks requested from Providers because they cannot be removed by running a Provider without the Task. Instead, the Task needs to be manually flagged in a way so that the Director will remove it.

\subsubsection{Priority:}

Priority in the Director is determined based on the closest common ancestor of the two competing Tasks. For root tasks, the closest common ancestor will be the root element. Once the closest common ancestor is determined, the priority of each Task's branch will determine which Task has higher priority. The winner takes control and becomes active, while the evicted Task will watch for an opportunity to take back control.

When a Task's ancestor tree has an optional Task between itself and the common ancestor, it is considered optional. If one Task has an optional parentage and the other does not, then the optional Task will automatically lose. The Tasks are compared normally if both have optional Tasks in their parentage.

\begin{figure}
    \centering
    \includegraphics[width=\textwidth]{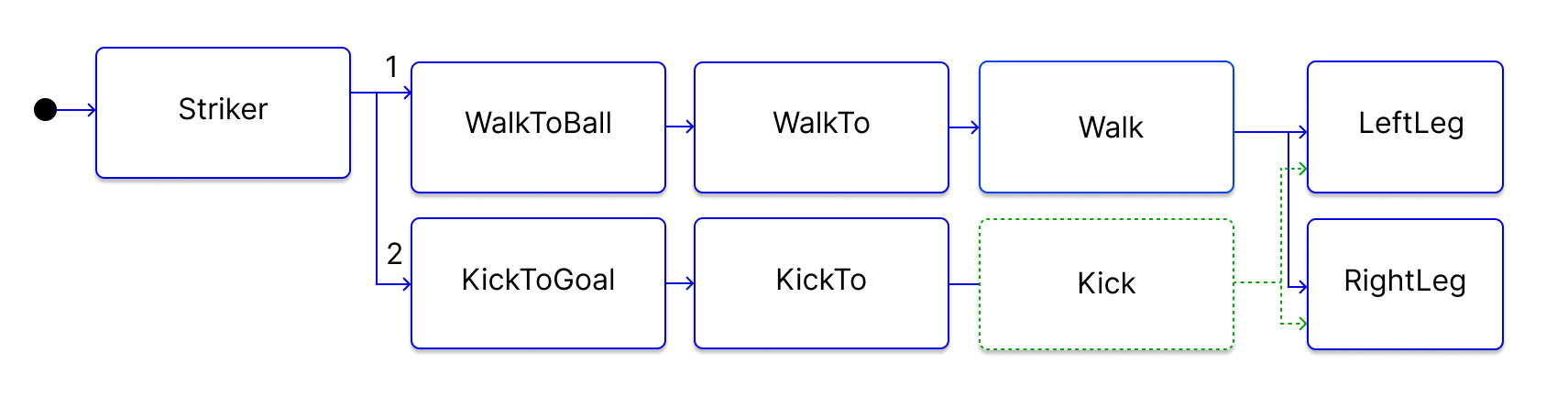}
    \caption{An example of priority, where the kick challenges the walk for control of the legs and succeeds in taking control due to its higher priority.}
    \label{fig:priority}
    \vspace{-2.5em}
\end{figure}

In Figure~\ref{fig:priority}, the walk is originally running and has priority over the left and right legs. The Kick then is requested and will attempt to take control of the legs. This becomes a challenge between the kick and walk Providers. The closest common ancestor of these two Providers is the striker Provider. Because KickToGoal has a higher priority than WalkToBall, the Kick Provider will take control of the legs from the Walk.

\subsubsection{Done Tasks:}

A Done Task is a Task sent by a Provider to notify its parent. The Provider group that initiated the Task will be triggered again upon receiving this signal from its child. The Done Provider's Task remains in the tree unless it's a root Task.

In Figure~\ref{fig:system}, when the LeftHipYaw Provider reaches the desired motor position, it sends a Done Task to its parent. Once all motors in LeftLeg send Done Tasks, LeftLeg itself sends a Done Task, allowing the next sequence of motor positions to run.

\subsubsection{Idle Tasks:}

An Idle Task is a Task a Provider requests to signal to continue running its previous Tasks. For example, in our scenario, the KickTo can send an Idle Task when it is kicking and does not want to re-request the Kick Task. If it is re-requested, the Kick may run from the beginning and not reach the end of its motion. Often, Done Tasks are used to check if a Provider should Idle or run new Tasks.

\subsection{Data}

Providers can access data about the local state of the Director to assist in making decisions and understanding the system state.

\subsubsection{Uses:}

`Uses' is a DSL keyword used to obtain information about subtasks. It provides information about the run state of the subtask and whether it is Done. The run states include whether the subtask is currently running, queued to run, or not yet requested by the current Provider. An example of how this information can be used is in Listing~\ref{code:kickto}

\begin{listing}[!t]
\begin{minted}[
frame=lines,
framesep=2mm,
baselinestretch=1.2,
fontsize=\footnotesize,
linenos
]{python}
def Provide(KickTo):
    # Kick is current running, keep running (idle)
    if kick.run_state is running and kick is not done:
        Task(Idle)
    if robot is aligned to target and ball in front of robot:
        # Need to kick and waiting to kick, keep waiting (idle)
        if kick.run_state is queued:
            Task(Idle)
        # Need to kick and we aren't trying to kick, send a new kick task
        else:
            Task(Kick)
    # Else no task, Director tree removes any existing Kick task
\end{minted}
\vspace{-1pt}
\caption{Pseudocode for the KickTo Provider using run state and done information.}
\label{code:kickto}
\end{listing}

\subsubsection{Run Reason:}

The `RunReason' DSL keyword retrieves information about why a Provider is running.

There are five possible run reasons - a new Task was requested, the Provider has become active, the Provider has become inactive, a subtask is Done, and the Provider is pushed because a higher priority module wants its Causing state to be true. The active and inactive RunReasons are relevant for the start and stop Provider types.

\section{Evaluation}\label{sec:discussion}

\subsection{Composability}

It is simple to combine modules in the Director. Requesting relevant Tasks combines desired modules into the program. Calling a `ChaseBall' Task from a non-Provider function or main loop of the program can easily demonstrate the robot chasing a ball. The Task call will trigger all the necessary modules to chase a ball through subtasks. The easy modification and addition of behaviours are critical for debugging and developing in high-pressure scenarios, such as RoboCup. Subsumption-based systems~\cite{Brooks1986} have more difficulty achieving this, as the layers build upon each other and are more tightly coupled.

\subsection{Extensibility}

In the Director system, adding new components is straightforward and does not require a global reevaluation of priorities. New Providers and Tasks can be created and requested without changes to the rest of the system. Subsections of the system can be easily isolated. This flexibility allows for easy experimentation, modification and debugging.

In our experience, the development of behaviour modules within the Director framework is significantly quicker than in our previous subsumption-like system. After converting our system, the high modularity made it easy to see ways to extend the functionality of the robots. It also allows for quick prototyping of research ideas, motivating the creation of an LLM strategy module with safety systems powered by the Director~\cite{OBrien2023}.

\subsection{Transitions}

In humanoid robotics, transitions are a factor in preventing the robot from causing damage to itself, others and the environment. The Director supports clean and safe transitions through conditionals on Providers to prevent them from running unless the system is in a particular state. Additionally, the Director includes functionality for soft transitions. Rather than immediately transitioning from one action to another, a Provider can handle the transition using the pushing functionality. These features facilitate high-priority safety modules critical for practical robotics.

In the previous section, we used a scenario to explain the soft transition functionality with the `When' and `Causing' keywords. The Kick Provider's `When' condition required the system stability state to be equal to `standing'. A matching `Causing' Walk Provider satisfied this state by making the walk engine stop cleanly. When the Kick Provider tried to take control, it pushed the Walk Provider group to use the `Causing' Walk Provider. The Walk then stops cleanly before transitioning to the Kick Provider. 

This smooth transition from walking to kicking is critical for stability. Conditions on Providers make the robot act safely and move between motions smoothly. Existing literature rarely addresses transitions, and we are not aware of other systems with soft transitioning functionality.

\subsection{Hardware Control}

The Director has a strict control system where a Provider can only provide for one Task at a time, and only one Provider of a Task type can be active at a time. Other Tasks and Providers are blocked and must wait for an opportunity to take control. This core rule in the algorithm prevents multiple sources from controlling one resource, such as the motors. By grouping motors, the system can ensure that only one module controls a kinematic chain. 

In our previous system, all motor commands moved through one module that enforced a similar strict control system. Another solution proposed by Bestmann and Zhang uses the Humanoid Control Module~\cite{Bestmann2020} to implement a mutex on the motors. These approaches lack the modularity and composability that the Director inherently creates throughout the system, from high-level strategy modules down to low-level motor modules. Instead of implementing locking mechanisms directly, the Director handles these tasks on behalf of the user.

\subsection{Versatility}

The Director is a versatile framework with extensive functionality. Research often focuses on adding one specific functionality. The Director aims to incorporate all needed functionality. Functionality for transitions and strict hardware control, as described previously, are critical parts of this versatility.

Another important aspect is the ability to create multiple implementations for one action. Provider groups facilitate this, where one implementation runs based on the system state. Numerous research articles address this concept in the context of the quality and existence of environment information~\cite{Veloso2000,Safronov2020,Yang2021}. Provider groups provide this functionality in a generalised way, where conditions determine the active Provider.

Modern implementations of behaviour trees for autonomous robotics request tasks and act based on the state of the subtask~\cite{Styrud2022,Safronov2020,Yang2021}. The Uses information and Done Tasks within the Director provide this functionality and could extend to conform to the behaviour tree structure if desired by adding fail and success responses from subtasks.

$ABC^2$~\cite{Matellan1998} has a similar Provider-Task relationship and can apply conditionals to nodes. They do not address transitions, resource control and multiple implementations for tasks. These are not as important within the two-dimensional simulation competition scenario used in the article but are important in the context of real-world humanoid robotics.

% \subsection{Transparency}

% Transparency aids debugging, but also results in more overhead and complexity to individual behaviours. Providers in the Director only have access to the system state through Task information and messages from non-Director modules, such as the vision system or localisation system. Having access to more information than is necessary for that individual behaviour is an anti-pattern in Director programming. The Director is designed to be decoupled, allowing for shorter and more composable modules without dependencies on other aspects of the system.

% The underlying algorithm driving the Director has a complete view of the system state, with all Providers and their active tasks and watchers visible at any given time. Using this information to create a graphical representation of the behaviour system is a valuable real-time debugging method.

\section{Conclusion}

The Director framework is a unique solution in the realm of behaviour systems for autonomous robotics. By addressing the limitations of existing frameworks, such as transitions and functionality, the Director offers a comprehensive solution for managing robot actions effectively and efficiently. It facilitates safety components and complex behaviours, critical for advanced humanoid robots in the real world.

The Director's ability to define multiple implementations for the same task based on conditionals adds a layer of flexibility and complexity. This feature enables agents to adapt their actions dynamically to changing environmental conditions, leading to more robust and versatile performance in real-world scenarios. The Director's strict resource control mechanisms, extending to motor control, ensure efficient and non-conflicting usage of resources.

The successful implementation of the Director in the RoboCup Humanoid Soccer League demonstrates its practical applicability and effectiveness in competitive environments. By simplifying the implementation of new behaviours and providing flexibility for integrating advanced technologies like LLMs, the Director opens up new possibilities for autonomous robotic behaviour.

% We presented the Director framework and placed it within the context of existing behaviour systems. It is modular, composable, extensible and has functionality critical for autonomous robotic systems. The Director supports soft transitions, multiple implementations for the same task chosen based on conditionals, conditional requirements on Providers, and strict resource control extending to motor control. 

% moved to top for space
\subsubsection*{Acknowledgements.} This research is supported by 4Tel Pty Ltd, 4AI Systems Inc and an Australian Government Research Training Program Scholarship to the first author. We acknowledge contributors of the NUbots Robotics Research Group, whose work this publication builds upon. Thank you to Johanne Montano for their review of this work.

% ---- Bibliography ----
%
% BibTeX users should specify bibliography style 'splncs04'.
% References will then be sorted and formatted in the correct style.
%
\bibliographystyle{splncs04}
\bibliography{nubots}
%
% \begin{thebibliography}{8}
% \bibitem{ref_article1}
% Author, F.: Article title. Journal \textbf{2}(5), 99--110 (2016)

% \bibitem{ref_lncs1}
% Author, F., Author, S.: Title of a proceedings paper. In: Editor,
% F., Editor, S. (eds.) CONFERENCE 2016, LNCS, vol. 9999, pp. 1--13.
% Springer, Heidelberg (2016). \doi{10.10007/1234567890}

% \bibitem{ref_book1}
% Author, F., Author, S., Author, T.: Book title. 2nd edn. Publisher,
% Location (1999)

% \bibitem{ref_proc1}
% Author, A.-B.: Contribution title. In: 9th International Proceedings
% on Proceedings, pp. 1--2. Publisher, Location (2010)

% \bibitem{ref_url1}
% LNCS Homepage, \url{http://www.springer.com/lncs}. Last accessed 4
% Oct 2017
% \end{thebibliography}
\end{document}